\documentclass[10pt,twocolumn,letterpaper]{article}

\usepackage[pagenumbers]{cvpr} 

\definecolor{cvprblue}{rgb}{0.21,0.49,0.74}
\usepackage[pagebackref,breaklinks,colorlinks,allcolors=cvprblue]{hyperref}

\def\paperID{*****} 
\def\confName{CVPR}
\def\confYear{2025}

\title{A{\normalsize RINAR}: Bi-Level Autoregressive Feature-by-Feature Generative Models}

\author{Qinyu Zhao \quad Stephen Gould \quad Liang Zheng\\
Australian National University \\
{\tt\small \{qinyu.zhao, stephen.gould, liang.zheng\}@anu.edu.au}
}

\begin{document}
\maketitle

\begin{abstract}
Existing autoregressive (AR) image generative models use a token-by-token generation schema. That is, they predict a per-token probability distribution and sample the next token from that distribution. The main challenge is how to model the complex distribution of high-dimensional tokens. Previous methods either are too simplistic to fit the distribution or result in slow generation speed. Instead of fitting the distribution of the whole tokens, we explore using a AR model to generate each token in a feature-by-feature way, \ie, taking the generated features as input and generating the next feature. Based on that, we propose A{\small RINAR} (AR-in-AR), a bi-level AR model. The outer AR layer take previous tokens as input, predicts a condition vector $\boldsymbol{z}$ for the next token. The inner layer, conditional on $\boldsymbol{z}$, generates features of the next token autoregressively. In this way, the inner layer only needs to model the distribution of a single feature, for example, using a simple Gaussian Mixture Model. On the ImageNet 256$\times$256 image generation task, A{\small RINAR}-B with 213M parameters achieves an FID of 2.75, which is comparable to the state-of-the-art MAR-B model (FID=2.31), while five times faster than the latter. Our code is available at \href{https://github.com/Qinyu-Allen-Zhao/Arinar}{https://github.com/Qinyu-Allen-Zhao/Arinar}.
\end{abstract}

\section{Introduction}
\label{sec:intro}
Autoregressive (AR) models are challenging diffusion models in the image generation task. Existing AR models use a token-by-token prediction schema. During training, an autoencoder converts training images into tokens, for example, 256 16-dimensional tokens for each image. An AR model uses tokens already generated as input and predicts the distribution of the next token. At inference time, the AR model keeps predicting the distribution of the next token and sampling a token from the distribution.

One of the main challenges is how to model the complex token distribution. A common practice is to quantize the continuous tokens into discrete codes and use multinomial distribution for sampling tokens. This introduces quantization errors, damaging the model performance. Another trend is to directly model the distribution of continuous tokens. GIVT~\citep{tschannen2023givt} predicts the parameters of a Gaussian Mixture Model (GMM) to model the token distribution and sample new tokens from the GMM. Instead, MAR~\citep{li2025mar} uses a lightweight diffusion model to generate tokens from noise, which implicitly learns the token distribution. Due to the limited expressive ability of GMM, GIVT is suboptimal to to fit the token distribution, while MAR is slower because of the diffusion procedure.

\begin{figure}[tb]
  \centering
   \includegraphics[width=0.98\linewidth,trim=0 100 300 0,clip]{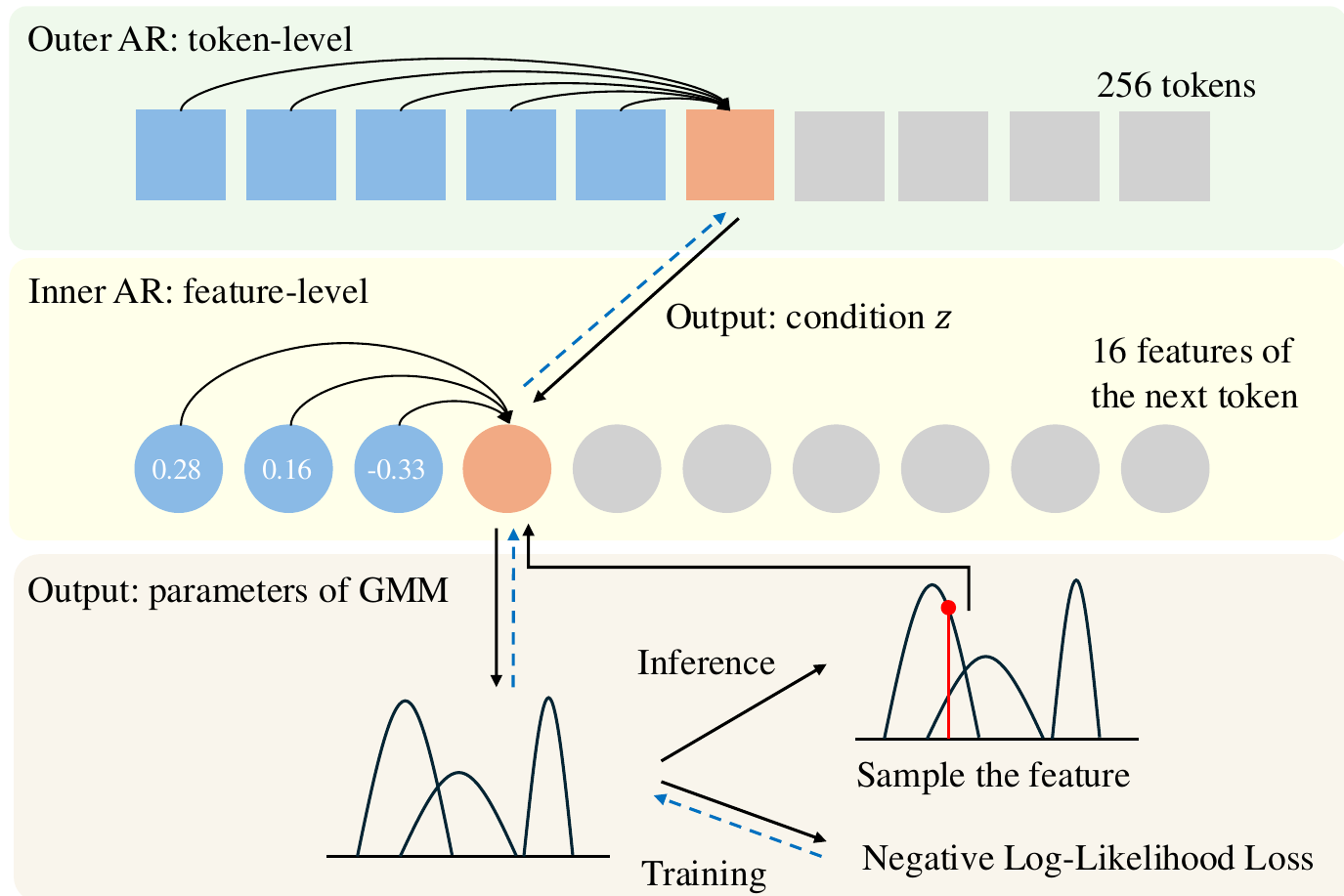}
   \caption{\textbf{Design of A{\small RINAR}}. The model consists of two AR layers. The outer layer predicts a condition vector $\boldsymbol{z}$, while the inner layer autoregressively generates the features of the next token based on $\boldsymbol{z}$. The blue dashed arrows indicate the gradient flow. GMM means the Gaussian Mixture Model. }
   \label{fig:method}
\end{figure}

In this report, we explores an AR feature-by-feature generation method. Instead of predicting the distribution of the whole token, the model predicts features of the next token autoregressively. That is, when generating a value for the $i$th feature of the token, the model takes the preceding features as input, \emph{i.e.}, features $1,2,...,i-1$. Based on this, we propose a bi-level AR model, A{\small RINAR} (AR-in-AR).

As shown in \cref{fig:method}, A{\small RINAR} consists of two autoregressive layers. The outer layer works on the token level. It takes tokens previously generated as input and predicts a condition vector $\boldsymbol{z}$ for the inner layer. The inner layer, conditional on $\boldsymbol{z}$, further generates each feature of the token autoregressively. The whole token generated by the inner layer will be fed back into the outer layer. Assume that an image is embedded into 256 16-dimensional tokens. The outer layer runs 256 times to generate condition vectors for tokens. Each time when generating a token, the inner layer will run 16 times to generate each feature of the token.

A concurrent work, fractal generative model~\citep{li2025fractal} proposes a similar idea. We claim that we independently get the design and want to highlight some differences. First, their models directly work on the pixel values of images, our model uses a latent representation from an autoencoder and generates images in a feature-by-feature method. Second, A{\small RINAR} achieves significant better performance and faster generation speed than their models.

We evaluate our model on the ImageNet 256$\times$256 image generation task. We train a base model with 213M parameters for 400 epochs and achieve a Fréchet Inception Distance (FID) score of 9.17 without CFG and a FID of 2.75 with CFG. With the same model size, it achieves performance comparable to the state-of-the-art MAR-B model (FID=2.31), while 5$\times$ faster than the latter.

To sum up, this report introduces an autogressive feature-by-feature generative model. We train a base model on ImageNet 256$\times$256 and achieve competitive performance. Our report indicates the great potential of pure autoregressive model on image generation.

\section{Related Works}\label{sec:relate_works}
\textbf{Autoregressive image generation.} AR models are popular in natural language processing and, recently, show good potential for image generation~\citep{esser2021taming, ramesh2021zero}. Most AR generative models use a tokenizer, \eg, VQVAE~\citep{razavi2019generating}, which converts images to discrete tokens. The AR model is trained to predict the discrete tokens with the cross-entropy loss. When generating images, the model predicts the distribution of the next token and then samples the next token autoregressively. The generated sequence of tokens are fed to tokenizer to obtain the final image. \citet{tian2025var} propose a new AR image generation schema, next-scale prediction, and their model surpasses the performance of diffusion models, indicating the competitive potential of AR generative models.

\textbf{Modeling continuous tokens.} Recent works propose to use continuous tokens instead of discrete ones to avoid quantization errors, where challenges are how to model the distribution of continuous tokens and how to sample from the distribution~\citep{li2025mar}. GIVT~\citep{tschannen2023givt} uses GMM to model the token distribution. When the model generates the next token, it predicts the parameters (mean, covariance, and weights) of a GMM and then samples the next token from the GMM. The model is trained with negative log-likelihood loss. MAR~\citep{li2025mar} uses a lightweight diffusion model head to fit the token distribution. When the AR model generates a condition vector $\boldsymbol{z}$ for the next token, the diffusion head will generate the next token from noisy data, conditional on $\boldsymbol{z}$. The model is trained with the loss of diffusion models.

All these works try to learn and predict the distribution of the entire token vectors. In comparison, our method at its inner layer learns and predicts the distribution of single features in a token.

\textbf{Fractal generative models.} A concurrent work, FractalMAR~\citep{li2025fractal} shares a similar idea. They propose a multi-layer autoregressive model, \eg, with 4 layers. The outermost model predicts the next patch, the second layer predicts the next region within the patch, the third layer predicts the next pixel within the region, and the innermost layer predicts the RGB channels within the pixel. Each outer layer predicts a condition vector for the inner layer, while the innermost layer predicts the distribution of RGB values of one pixel.

We claim that we independently designed our A{\small RINAR}. Besides, FractalMAR works on the pixel space, generating an image pixel by pixel, while A{\small RINAR} uses a feature-by-feature method and achieves better performance and faster speed. Essentially, A{\small RINAR} can be considered as a latent-space version of FractalMAR.

\section{Methods}\label{sec:method}
\subsection{Model Architecture}
First, we use an autoencoder, converting images to continuous tokens, following \citet{li2025mar}. Here, each image is converted to 256 16-dimensional tokens.

As shown in \cref{fig:method}, A{\small RINAR} consists of two layers of AR models. The outer layer has much more parameters, while the inner layers are smaller. The outer layer takes tokens  generated previously as input, and generates a condition vector $\boldsymbol{z}$ for the next token. $\boldsymbol{z}$ is fed into the inner layer as a condition. 

Then, the inner layer generates the features of the next token autoregressively. Specifically, the inner layer takes the generated features as input and predicts the parameters of a GMM as the distribution of the next feature, conditional on $\boldsymbol{z}$. The next feature is sampled from the GMM, and the sampled feature is fed into the inner layer again, to obtain the distribution of the following features. After the inner layer generates all features of the token, the outer layer takes the generated token as input and predicts the condition vector for the following token. The model continues this process until it has generated all tokens of the image.

Note that our design is a general architecture. In this paper, we use MAR~\citep{li2025mar} as the outer AR layer and a causal transformer with  adaptive normalization (AdaLN) as the inner layer. But these two layers can be replaced with other existing AR model architectures.

\begin{table*}[ht]
    \centering
    \caption{\textbf{System-level method comparison} on ImageNet 256×256 conditional generation. With a similar model size, the performance of A{\small RINAR} is comparable to the SoTA MAR-B model and is superior to FractalMAR-B.}
    \label{tab:comparison}
    \begin{tabular}{l l | c | c c | c c | c c | c c}
        \toprule
          & &   &  \multicolumn{4}{c|}{ w/o CFG} & \multicolumn{4}{c}{w/ CFG} \\
        &  & \#params  &  {FID$\downarrow$} & {IS$\uparrow$} & {Pre.$\uparrow$} & Rec.$\uparrow$ & {FID$\downarrow$} & {IS$\uparrow$} & {Pre.$\uparrow$} & {Rec.$\uparrow$} \\
        \midrule
        \multicolumn{2}{l}{\hspace{-.5em} \textit{pixel-based}} \\
        & ADM~\citep{Dhariwal2021} & 554M & 10.94 & 101.0 & 0.69 & 0.63 & 4.59 & 186.7 & 0.82 & 0.52 \\
        & VDM$++$~\citep{Kingma2023} & 2B & 2.40 & 225.3 & - & - & 2.12 & 267.7 & - & - \\
        & FractalMAR-B~\citep{li2025fractal} & 186M & 137.4 & 8.8 & 0.13 & 0.31 & 11.80 & 274.3 & 0.78 & 0.29 \\
        \midrule
        \multicolumn{2}{l}{\hspace{-.5em} \textit{vector-quantized tokens}}   \\
        & Autoreg. w/ VQGAN~\citep{Esser2021} & 1.4B & 15.78 & 78.3 & - & - & - & - & - & - \\
        & MaskGIT~\citep{Chang2022} & 227M & 6.18 & 182.1 & 0.80 & 0.51 & - & - & - & - \\
        & MAGE~\citep{Li2023} & 230M & 6.93 & 195.8 & - & - & - & - & - & - \\
        & MAGVIT-v2~\citep{Yu2024} & 307M & 3.65 & 200.5 & - & - & 1.78 & 319.4 & - & - \\
        & VAR-d16~\citep{tian2025var} & 310M & - & - & - & - & 3.30 & 274.4 & 0.84 & 0.51 \\
        \midrule
        \multicolumn{2}{l}{\hspace{-.5em} \textit{continuous-valued tokens}}  \\
        & LDM-4$^\dagger$ \citep{Rombach2022} & 400M & 10.56 & 103.5 & 0.71 & 0.62 & 3.60 & 247.7 & 0.87 & 0.48 \\
        & U-ViT-H/2-G~\citep{bao2022all} & 501M & - & - & - & - & 2.29 & 263.9 & 0.82 & 0.57 \\
        & DiT-XL/2~\citep{Peebles2023} & 675M & 9.62 & 121.5 & 0.67 & 0.67 & 2.27 & 278.2 & 0.83 & 0.57 \\
        & DiffT~\citep{hatamizadeh2023diffit} & - & - & - & - & -  & 1.73 & 276.5 & 0.80 & 0.62 \\
        & MDTv2-XL/2~\citep{Gao2023} & 676M & 5.06 & 155.6 & 0.72 & 0.66 & 1.58 & 314.7 & 0.79 & 0.65 \\
        & GIVT~\citep{tschannen2023givt} & 304M & 5.67 & - & 0.75 & 0.59 & 3.35 & - & 0.84 & 0.53 \\
        & MAR-B~\citep{li2025mar} & 208M & 3.48 & 192.4 & 0.78 & 0.58 & 2.31 & 281.7 & 0.82 & 0.57 \\
        \midrule
        & A{\small RINAR}-B ($t=1.0$) & 213M & 9.17 & 123.9 & 0.70 & 0.61 & 2.75 & 278.3 & 0.82 & 0.55 \\
        & A{\small RINAR}-B ($t=1.1$) & 213M & 6.28 & 145.8 & 0.74 & 0.58 & 3.13 & 265.2 & 0.82 & 0.53 \\
        \bottomrule
    \end{tabular}
\end{table*}
\begin{table}
    \centering
    \caption{\textbf{Method comparison on generation efficiency}. A{\small RINAR} is significantly faster than previous works, while maintaining competitive generation quality. }
    \label{tab:speed}
    \begin{tabular}{l  | c | c}
        \toprule
         Model & FID & Time / image (s) \\
        \midrule
        MAR-B~\citep{li2025mar} & 2.31 & 65.69 \\
        FractalMAR-B~\citep{li2025fractal} & 11.80 & 137.62 \\
        A{\small RINAR}-B & 2.75 & 11.57 \\
        \bottomrule
    \end{tabular}
\end{table}

\subsection{Temperature and Classfier-Free Guidance}
Following GIVT~\citep{tschannen2023givt}, we implement temperature sampling for A{\small RINAR}. In short, when A{\small RINAR} predicts a GMM for a feature, the standard deviation of Gaussians in GMM will be divided by a temperature parameter. When temperature is greater than 1, the distributions will be shrinked and sharper to the mean. When temperature is less than 1, the distributions will be flattened.

Besides, we implement CFG for A{\small RINAR}. The model is trained with an additional null class to learn the unconditional distribution. At inference time, the model is run twice, the first time with class condition, while in the second time, without class condition. We subtract the unconditional probabilities from the class-conditional probabilities, and then sample the next features. With a good temperature and CFG, model performance can be further improved.

\begin{figure*}
  \centering
   \includegraphics[width=0.99\linewidth,trim=0 120 0 0,clip]{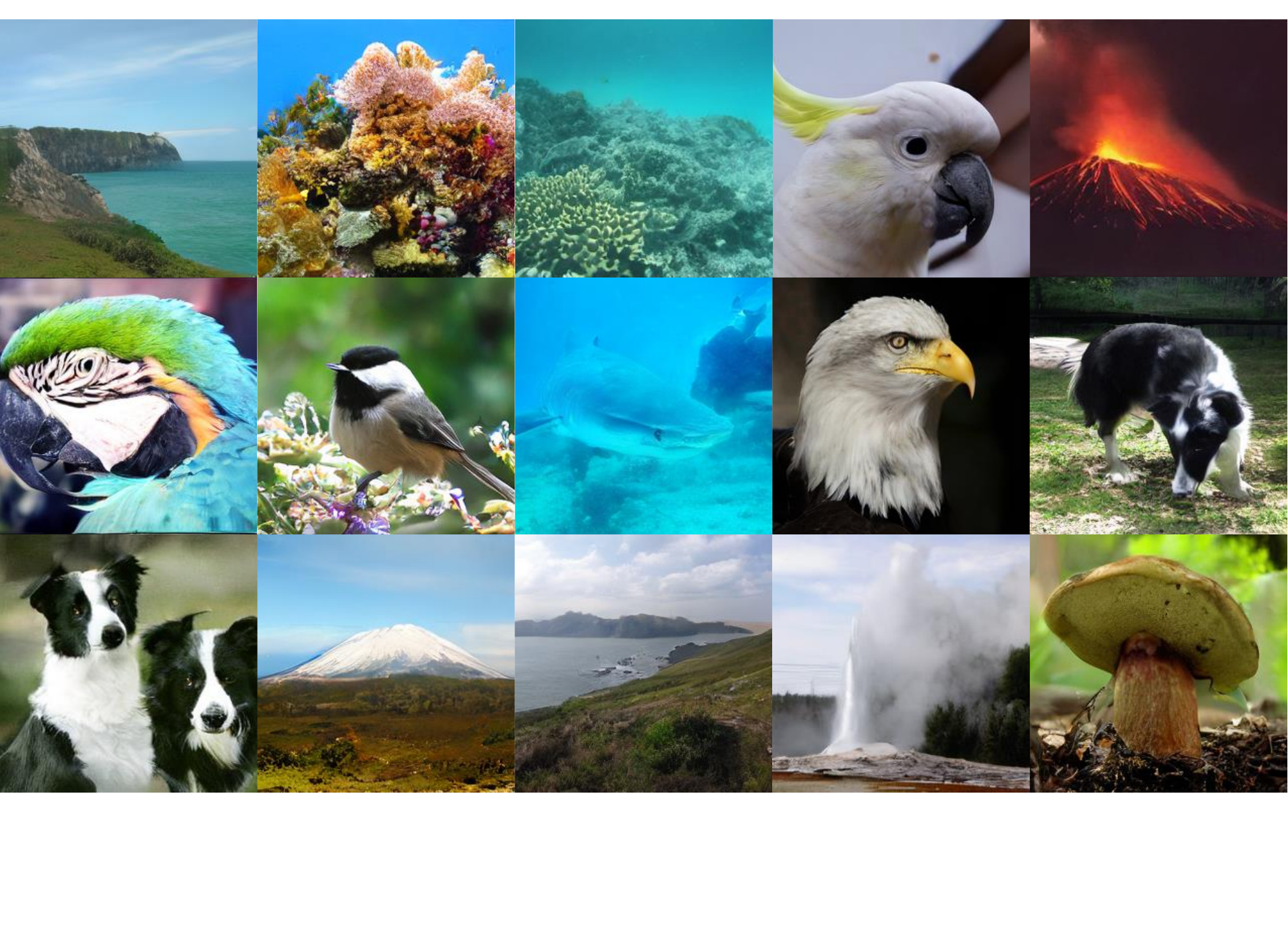}
   \caption{\textbf{Qualitative Results.} We show selected examples of class-conditional generation on ImageNet 256$\times$256 using A{\small RINAR}-B.}
   \label{fig:examples}
\end{figure*}

\section{Experiments}\label{sec:experiments}
\subsection{Implementation Details}
We follow the training recipe of MAR~\citep{li2025mar}. Specifically, a 213M model, A{\small RINAR}-B, is trained on the ImageNet at a resolution of 256$\times$256 for 400 epochs.

We conduct a preliminary experiment to decide the number of Gaussians in GMM for modeling the feature distribution. Specifically, we use a MAR-B model pretrained by \citet{li2025mar} as the outer layer, which is frozen in the following procedure. Then, we initialize and train the inner layer on ImageNet for 3 epochs and evaluate the model. When the number of Gaussians that the model head predicts is changed, we observe that small number of Gaussians (less than 3) are too simplilistic to fit the feature distirbution, while more than 4 Guassians do not show more benefit. Thus, we use 4 Gaussians to model the feature distribution in the inner AR layer.

All experiments are run on 4 NVIDIA A100 GPUs, and it takes about 8 days to train A{\small RINAR}-B. Details can be found in the supplementary materials. 

\subsection{Conditional Image Generation}
To evaluate the proposed model, we use A{\small RINAR}-B to generate 50K images with class as condition. We then calculate FID~\citep{fid} and IS~\citep{is}, and also provide Precision and Recall as references. We follow the evaluation suite provided by \citet{Dhariwal2021}.

We compare our model with existing generative models. Results are shown in \cref{tab:comparison}. As seen, our 213M model achieves a FID of 9.17 without CFG, which is comparable to a larger diffusion model DiT-XL/2. With CFG, the performance of our model is improved to FID=2.75. Note that our model outperforms FractalMAR-B~\citep{li2025fractal} and is comparable to the SoTA MAR model. These results indicate the potential of feature-by-feature image generation.

An interesting observation is that temperature $t=1.1$ improves the model performance without CFG but deteriorates it with CFG. We hypothesize that either a larger temperature or CFG sharpens the predicted distributions, so features would be sampled in a smaller region, reducing outlier generations. But applying the two techniques together may lose generation diversity.

Sample images generated by A{\small RINAR}-B are shown in \cref{fig:examples}. We provide more examples and failure cases in the supplementary materials.

\subsection{Generation Speed}
An important aspect of generative models is efficiency to meet user needs. 
We evaluate the generation efficiency of A{\small RINAR}-B, FractalMAR-B~\citep{li2025fractal}, and MAR-B~\citep{li2025mar} in \cref{tab:speed}. All models are run on a single NVIDIA A100 GPU (80G) and generate 100 images one by one. The total running time is averaged to obtain the generation time per image. Results show that our model is significantly faster than the compared works while maintaining a good FID. This is probably due to the use of the Gaussian mixture modeling of feature distributions instead of using a diffusion model.

\section{Conclusion}
In this report, we introduce a feature-by-feature generative model. Our model has two AR layers: the outer layer generates tokens sequentially, and the inner layer generates the 16 dimensions of a single token autoregressively. We test our model on ImageNet 256×256, using a 213M parameter base model. The model performance is comparable to the state of the art while speeding up the generation process. Our results show that our approach is effective for generating high-quality images. We will conduct more ablation and variant studies and scale up this model in our report updates.

{
    \small
    \bibliographystyle{ieeenat_fullname}
    \bibliography{main}
}

\clearpage
\setcounter{page}{1}
\maketitlesupplementary

\section{Implementation Details}
\label{supp_sec:details}
We provide the training hyperparameters in \cref{tab:training_hyperparameters} and model architectures in \cref{tab:model_architecture}

\begin{table}[h]
    \centering
    \caption{\textbf{Training Hyperparameters}}
    \begin{tabular}{l|l}
        \toprule
        Hyperparameter & Value \\
        \midrule
        Optimizer & AdamW~\citep{kingma2014adam,loshchilov2017decoupled} \\
        Weight decay & 0.02 \\
        AdamW moment & (0.9, 0.95) \\
        Total training epochs & 400 \\
        Warm-up epochs & 100 (linear LR warmup)~\citep{goyal2017accurate} \\
        Batch size & 256 \\
        Learning rate (LR) & $1 \times 10^{-4}$ \\
        LR schedule & constant \\
        \bottomrule
    \end{tabular}
    \label{tab:training_hyperparameters}
\end{table}

\begin{table}[h]
    \centering
    \caption{\textbf{Model Architectures.}}
    \begin{tabular}{l|l l}
        \toprule
        Hyperparameter & Outer AR & Inner AR \\
        \midrule
        Model & MAR-B~\citep{li2025mar} & AdaLN \\
        \#Parameters & 201M & 12M \\
        \#Blocks & 24 & 1 \\
        Width & 768 & 768 \\
        AR steps & 256 & 16 \\
        \bottomrule
    \end{tabular}
    \label{tab:model_architecture}
\end{table}

\section{Visualization of Generated Samples}
\label{supp_sec:examples}
\begin{figure}[h]
  \centering
   \includegraphics[width=0.99\linewidth,trim=0 520 120 0,clip]{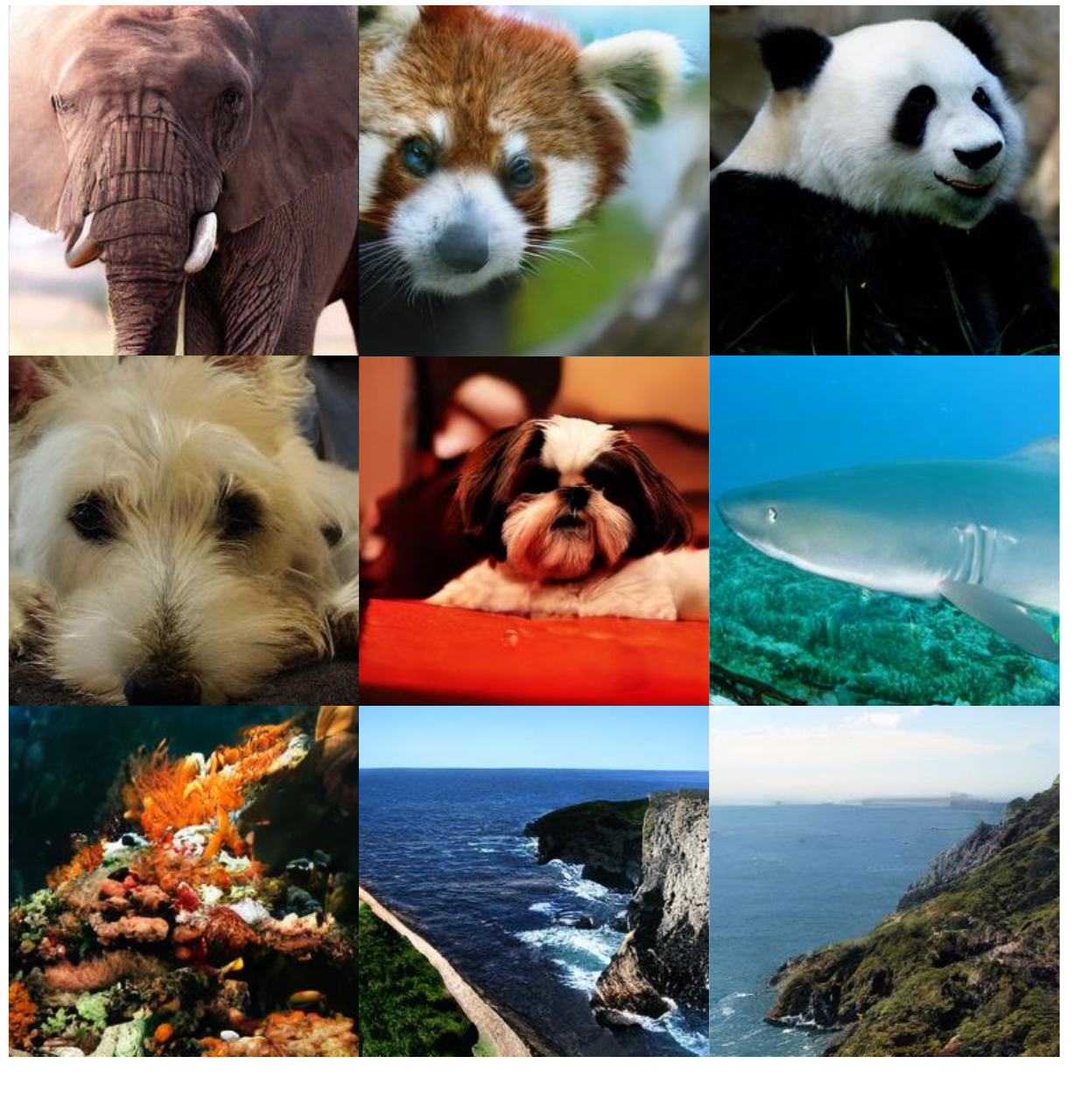}
   \caption{\textbf{Generated samples from A{\small RINAR}.}}
   \label{fig:extra_examples}
\end{figure}

\begin{figure}[h]
  \centering
   \includegraphics[width=0.99\linewidth,trim=0 120 370 0,clip]{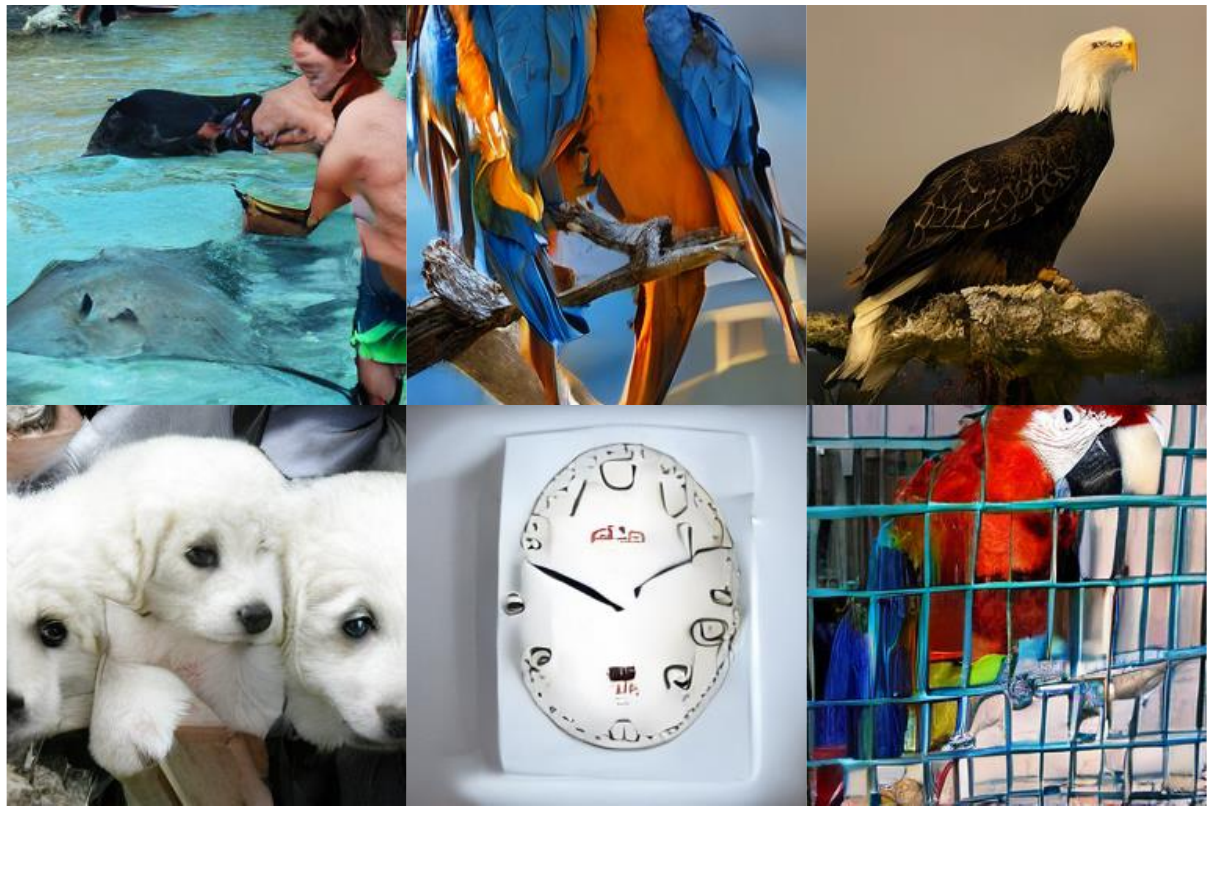}
   \caption{\textbf{Failure cases.} Some cases are very challenging to A{\small RINAR}, such as human bodies, multiple objects, text, and complex structure. The generated images contain noticeable artifacts.}
   \label{fig:failure}
\end{figure}

Additional visualization results generated by A{\small RINAR} are provided in \cref{fig:extra_examples} and some failure cases are shown in \cref{fig:failure}.

\end{document}